# Challenges of YOLO-Series for Object Detection in Extremely Heavy Rain: CALRA Simulator-based Synthetic Evaluation Dataset

Taesoo Kim, Hyeonjae Jeon, and Yongseob Lim[†], *Member, IEEE*

*Abstract—* Recently, as many studies of autonomous vehicles have been achieved for levels 4 and 5, there has been also increasing interest in the advancement of perception, decision, and control technologies, which are the three major aspects of autonomous vehicles. As for the perception technologies achieving reliable maneuvering of autonomous vehicles, object detection by using diverse sensors (e.g., LiDAR, radar, and camera) should be prioritized. These sensors require to detect objects accurately and quickly in diverse weather conditions, but they tend to have challenges to consistently detect objects in bad weather conditions with rain, snow, or fog. Thus, in this study, based on the experimentally obtained raindrop data from precipitation conditions, we constructed a novel dataset that could test diverse network model in various precipitation conditions through the CARLA simulator. Consequently, based on our novel dataset, YOLO series, a one-stage-detector, was used to quantitatively verify how much object detection performance could be decreased under various precipitation conditions from normal to extreme heavy rain situations.

## I. INTRODUCTION

In recent years, thanks to the development of various deep learning technologies and sensors, the autonomous driving level could be rapidly developed to 4 or 5 levels. [1-3]. As the level of autonomous driving technology has risen significantly, autonomous driving has become possible on most roads, especially in complex cities, curves, and alleys, making it easier to overcome unexpected situations. However, despite the advancement of autonomous driving technology toward level 4 or higher, driver intervention could be essential in autonomous driving in adverse weather conditions such as rain, snow, and fog [4].

Specifically, in order to successfully perform path planning and tracking control in autonomous driving, it could be very important to accurately recognize the surrounding situation even in adverse weather conditions of the autonomous vehicle through various sensors (LiDAR, radar, camera). However, as shown in Fig. 1, camera-based object detection performance for autonomous driving could be much more deteriorate in bad weather condition (e.g., heavy rain) rather than normal condition. Unlike radar and LiDAR which use ultrasonic waves and lasers respectively, camera sensors, which receive data through visible light, are indirectly blocked by rain, snow, and fog. In addition, raindrops or snow passing close to the camera can blur the background of the image due to a focus problem that goes beyond simply obscuring the object. Although the camera sensor is inexpensive and easy to use compared to other sensors (e.g., LiDAR and radar), it has the crucial disadvantage that it is be very vulnerable to bad weather conditions [5-8].

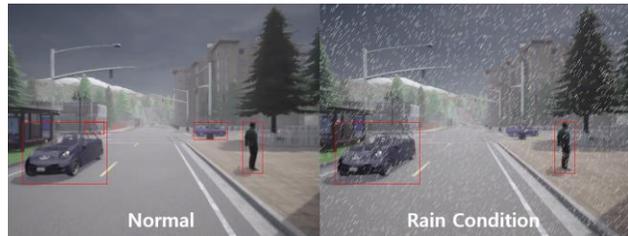

Figure 1. Performance degradation using camera-based object detection in rain condition.

To tackle these problems, many researchers have been conducted to improve object detection performance through cameras even in bad weather conditions by using deep learning networks. Thus, two major researches through deep learning based methods have been conducted to solve these problems.

First, they put many efforts to develop a model that could remove noises from videos and images. Then, this model which was able to obtain a clear image from a noisy image achieved good results in terms of object detection performance in the rain conditions. For example, DerainNet [9] effectively performed deraining on a single image by training the network on the detail layer of the image. In other studies, RESCAN [10] was proposed, and noisy images, in this study, were carefully analyzed using different alpha-values according to intensity and transparency to remove rain from the original image. In addition, PreNet [11] devised a progressive ResNet that repeatedly deployed a general ResNet and performed well in both synthetic and real rainy images. And, Restormer [12] proposed an efficient transformer model through a multi-head attention and generated forward network, and consequently achieved state of the art on the Test1200 dataset, achieving 33.19 PSNR with high accuracy in deblurring and deraining from original image in rain conditions. However, this method has the critical disadvantage that it is difficult to use in real world environments because the model is very heavy and slow to

*Research supported by the Daegu Gyeongbuk Institute of Science and Technology (DGIST), the National Research Foundation of Korea (NRF) grant funded by the Korean government (MSIT) from the Ministry of Science and Information and Communications Technology (ICT) of Korea.

Taesoo Kim, Hyeonjae Jeon, Yongseob Lim[†] are with DGIST, 42988 Daegu, Republic of Korea. (e-mail: xotn8888@dgist.ac.kr; wjsguswo12@dgist.ac.kr; yslim73@dgist.ac.kr.) †: Corresponding author.

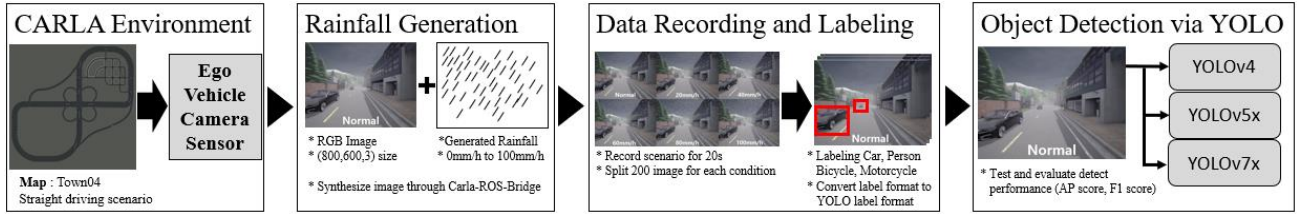
Figure 2. Overall workflow for virtual data preparation and object detection performance evaluation of YOLO-series.

use in self-driving cars that require detecting objects in real time.

Second, the research has been conducted to train real time object detectors such as YOLO (You only look once) and Faster-RCNN through synthetic images so that they can quickly detect objects even in noisy situations [13]. However, these methods heavily relied on the dataset used for learning, making it difficult to apply it to a new environment which was not seen in the dataset, and even collecting the appropriate dataset was also very difficult in terms of cost and time. For example, models
learned from images of 10mm/h precipitation situations or synthetic images was able to reveal low accuracy when the model faced with precipitation situations greater than 10mm/h. Thus, they also faced the data-hungry problem because they did not have much access to precipitation situations above 10mm/h.

Therefore, to solve this data-hungry problem which is very difficult to obtain real world data, realistic data can be created using deep learning techniques such as GAN in real images [14]. Some studies also created synthetic image with rain through mathematical calculations of raindrop numbers and Gaussian filters [15]. In addition, by synthesizing raindrops in real images, sim to real researches were recently being conducted to apply for the real world by creating datasets or learning deep learning models using data from the simulation environment models. Moreover, according to this study [16], sim to real domain adaptation was successfully performed by obtaining a kinetic image for lane recognition through CARLA. Thus, virtual environments such as CARLA [17], were already used by several researchers and were gaining high reliability for the bad weather conditions.

The crucial problem with the previous studies was revealed that the observed or synthesized precipitation was fixed to a specific value or was not sufficiently explained. Therefore, the object detection performance decreases due to rain, but how much precipitation decreases the performance has not been quantitatively studied. Focusing on the current problem of object detection and the factors of virtual simulation, this paper aims to evaluate the performance according to diverse rainfall rate for existing deep learning based real-time object detectors (i.e., Yolo-series) using CARLA environment and synthetic rain image of road driving situations as shown in Fig. 2. Therefore, this paper has the following original contributions.

- Based on the pre-trained deep learning models (i.e., YOLOv4, YOLOv5, YOLOv7) through coco datasets, we suggested the reliability of the novel virtual datasets constructed for this study was secured by comparing the test results using the KITTI dataset [18] in driving conditions without precipitation.

- Using various precipitation obtained based on precipitation experimental data, we compared and verified in great detail how much the performance of deep learning-based real-time object detection algorithms (YOLOv4, YOLOv5, YOLOv7) for autonomous driving was decreased under virtual precipitation conditions on CARLA environment.

- Moreover, we also discussed and suggested various network model development methods for improving the deep learning-based object detection performance under various precipitation conditions.

The remainder of paper is consisted of Section II to Section V. Section II introduces the object detection model and CARLA virtual environment related to this study. Section III explains the object detection model of applying precipitation evaluation method to CARLA simulation. Section IV explains the results of the model degradation due to precipitation, and Section V suggests an improvement method based on the results.

## II. RELATED WORK

### A. Real Time Object Detect Model

YOLO (You only look once) and Faster-RCNN have been studied as real-time object detection models. Faster-RCNN is a two-stage-detector model with good speed and performance, and YOLO is a one-stage-detector model with better speed than performance. Therefore, YOLO has been developed in the direction of increasing the accuracy while Faster-RCNN has been developed in the direction of increasing the speed of inference. Then, in the case of YOLO, using trainable bag-of-freebies, the YOLOv7-E6E recorded 56.8 box AP and 36 FPS, and YOLOv7x outstandingly achieved the 53.1 box AP and 114 FPS. Consequently, these models successfully accomplished the state of the art of the current real-time object detection technology [19]. However, in previous works, YOLOv4 had reached the 43 box AP and 31 FPS, YOLOv5x had achieved the 47.3 box AP and 153 FPS [20,21].

### B. Object Detection in Bad Weather Conditions

Object detection methods under bad weather conditions for the autonomous driving vehicle were also being studied in various ways. For example, Hnewa et al. tested the performance of the existing object detection model for the Bdd100k dataset [22, 23] and the deraining techniques such as DerainNet and PreNet. However, these deraining methods showed insufficient performance when they were applied to the actual driving environment. Accordingly, researches have been conducted to improve object detection performance in bad weather environments through new domain adaptation models [24], or to make images for inference clearer and sharper [25], or to improve object detection performance through data augmentation and ensemble methods [26].

Some researchers have conducted a study comparing clear weather and rainy weather using general YOLO models [27,28]. Abdulghani et al. conducted a study comparing performance with one-stage detector in bad weather conditions using Fast-RCNN which was originally the two-stage detector. According to this study, when comparing the YOLOv4 model and the Faster-RCNN, YOLOv4 was substantially outperformed in terms of object detection performance [29].

*C. Rain Synthetic Dataset Obtained from Real World*

For this study with high fidelity in terms of object detection performance in adverse weather conditions, the test accuracy at default weather in CARLA simulation should be the similar accuracy of real world, and the condition of rain in the simulation should be also similar specific precipitation of real world. To this end, according to the previous study [30], for realistic precipitation images from real experiments, the simulation precipitation level was verified using four actual precipitation situations (e.g., 10mm/h, 15mm/h, 25mm/h, 30mm/h). Moreover, the fidelity of synthetic precipitation level data was investigated by using-structural similarity index measure (SSIM) [31] and peak signal-to-noise ratio (PSNR) [32] as described in equations (1) and (2) respectively. SSIM and PSNR are respectively the following (1) and (2).

$$\text{SSIM}(x,y) = \frac{(2\mu_x\mu_y + C_1)(2\sigma_{xy} + C_2)}{(\mu_x^2 + \mu_y^2 + C_1)(\sigma_x^2 + \sigma_y^2 + C_2)} \quad (1)$$

$$\text{PSNR} = 10 \log_{10}\left(\frac{\text{MAX}_I^2}{\text{MSE}}\right) \quad (2)$$

where mean intensities of *x* and *y* direction are notated $\mu_x$ and $\mu_y$ respectively, which are needed for luminance comparison. Also, standard deviations of *x* and *y* direction are $\sigma_x$ and $\sigma_y$ respectively, which are needed for contrast comparison. $\sigma_{xy}$ is correlation coefficient. $\text{MAX}_I$ is maximum possible value of the image. MSE is mean squared error. Based on [31], PSNR and SSIM provide a basis for people to judge similar structures of images while they have realistically physical meaning. Therefore, it means that the higher each value, the higher the similarity with the original image.

According to research [30], the image was randomly sprayed with an increase of 10 to 5000 raindrops in units of 10, and the number of raindrops satisfying the following conditions in equations (3) and (4) was confirmed for PSNR and SSIM compared to the actual rain image. In this study, through this experiment, a linear relationship between the number of raindrops for realistic description and rainfall rate was obtained.

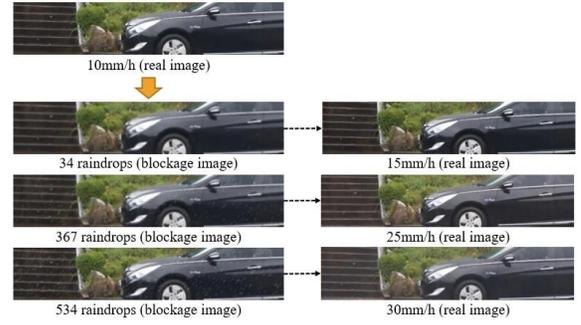

Figure 3. Comparing rain synthetic image (Left) and real raindrop image(right) [30].

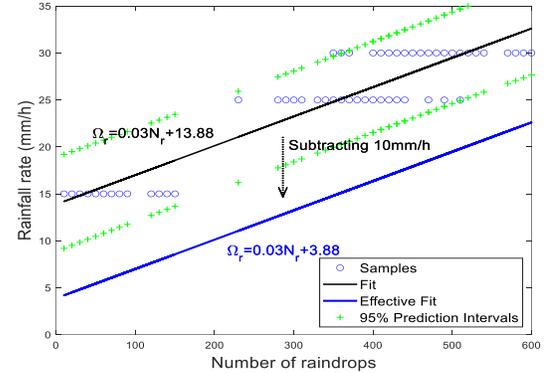

Figure 4. The relationship between number of raindrops and rainfall with 95% confidence level, R squared score of the regression results is 0.84 [30].

$$0.9500 < (SSIM) < 0.9510. \quad (3)$$

$$(PSNR) > 41.50. \quad (4)$$

As a result, in order to create a synthetic image similar to the actual precipitation environment, the number of raindrops as shown in Fig. 3 should be sprayed, and the correlation between the number of raindrops and rainfall rate can be expressed as shown in Fig. 4. In addition, when the range of the number of raindrops satisfying equations (3) and (4) is obtained, it can be expressed as the following equation (5).

$$\Omega_r = 0.03 \times N_r + 3.88 \quad (5)$$

where $\Omega_r$ is the rainfall rate in mm/h, $N_r$ is the number of raindrops in the image.

### III. DATA PREPARATION AND DETECTION VIA CARLA

*A. CARLA Rain Synthetic Image Preparation*

A dataset of general conditions without rain can be obtained from KITTI or Bdd100k. In particular, in the case of Bdd100k, various weather conditions such as cloudy weather, rainy weather, and fog are recorded in addition to general situations. In this study, we tried to collect data for evaluation through CARLA simulation because it was difficult to use existing datasets or collect data from 10mm/h to 100mm/h precipitation situations. To obtain test data, we used default map, Town04 of CARLA, as shown in Fig. 5. And spawn objects such as vehicles, bicycles, people, and motorcycle were shown in Fig. 6. The ego vehicle was driven straight to collect data in the CARLA environment for 20 seconds, and the images during driving were recorded through the proposed situation. The total number of objects placed on the map was shown in the Table I. The RGB image (800,600,3) was obtained from the RGB camera sensor mounted on the CARLA ego vehicle, which was transmitted to the Rviz via CARLA-ros-bridge, and then the rain image was synthesized. Finally, we obtained two video files: first one was the general driving video to be used as comparison group and second one was the rain synthesized video to be used as experimental group. The formula used to synthesize the rain image was as shown in equation (5). Also, driving scenario was recorded by changing rainfall rate from 0 mm/h to 100 mm/h in units of 10 mm/h. Finally, with a total of 11 videos, results of rain synthetic images in terms of rainfall rate were obtained, as shown in Fig. 7

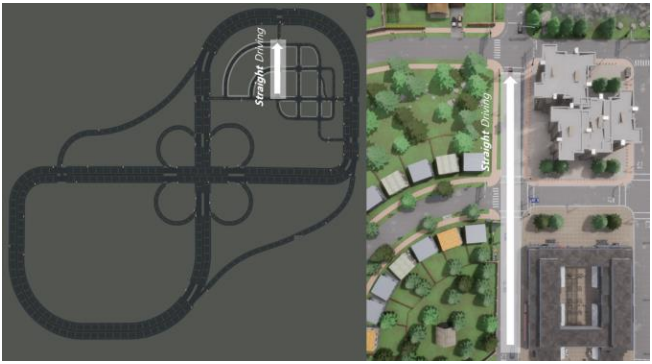

Figure 5. Scenario trajectory in CARLA default map, Town04 where the ego vehicle drove along the white arrow.

### B. Object Detection

After the video recorded in terms of rainfall rate was cut into frames, and we finally obtained a total of 200 images per video. As for these images used for test dataset, ground truth labeling was performed on the instances. At this time, the number of entire target instance is shown in the Table I. According to this study, YOLO was chosen as the model for object detection in diverse precipitation conditions because the YOLO model recorded high performance when comparing one-stage-detector YOLOv4 and YOLOv3 with two-stage-detector Faster-RCNN in bad weather conditions [29]. YOLOv7x, YOLOv5x, and YOLOv4 were used as models for evaluation, and each weight of YOLO-series was pretrained through COCO dataset. The test was conducted with 32 batch sizes. In addition, in order to obtain the reliability of the virtual dataset, KITTI dataset was also tested with the same model, which is used to compare scores on Car and Person. For the prepared test dataset, we tried to compare the F1 score and AP score according to diverse rainfall rate, and it was expected that the scores would gradually decrease as the rainfall rate increased.

TABLE I. TOTAL OBJECT IN CARLA ENVIRONMENT.

| Total Object | Number of Object | Number of Instance |
|---|---|---|
| Car | 6 | 838 |
| Person | 8 | 350 |
| Bicycle | 2 | 67 |
| MotorCycle | 1 | 10 |

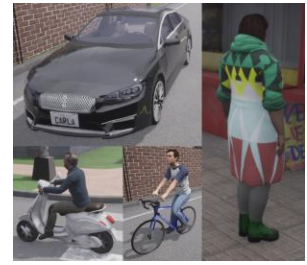

Figure 6. Used objects for test: car, person, bicycle, motorcycle.

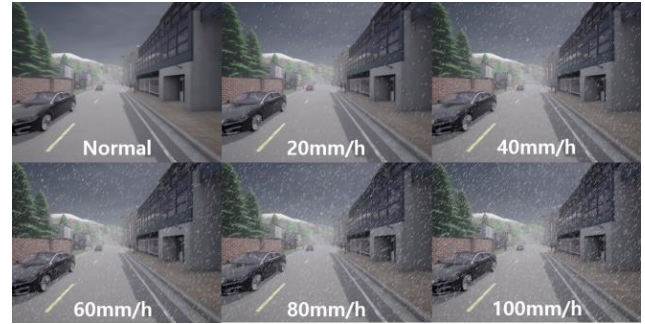

Figure 7. Example of rain synthetic images in terms of rainfall rate.

### IV. EXPERIMENT RESULTS

#### A. Comparison KITTI and CARLA Dataset

The 20-second driving video was divided into a total of 200 images including 838 cars, 350 people, 67 bicycles, and 10 motorcycle instances. KITTI dataset was a total of 7483 images including 4493 people, 2874 car instances. We tested YOLOv7x, YOLOv5x, and YOLOv4 for a total of 11 weather environments using normal weather of CARLA default map and images of rainfall rate from 10mm/h to 100mm/h.

To evaluate the reliability of the experiment in a virtual environment, each object detection model was tested on the KITTI road driving dataset. Table II shows the comparison of AP scores, F1 scores for KITTI, and scores in the default weather of CARLA. On average, AP scores for cars decreased by 0.152, and AP scores for person increased by 0.183 compared to KITTI dataset. Considering that CARLA data had a different domain and weather was cloudy rather than clear, it was determined that our data could have sufficient reliability in evaluating the object detection model.

TABLE II. COMPARISON KITTI AND CARLA DATASET ON NORMAL WEATHER CONDITION.

| Index values | | KITTI | | CARLA data | |
|---|---|---|---|---|---|
| | | *AP* | *F1* | *AP* | *F1* |
| YOLOv4 | Car | 0.805 | 0.555 | 0.64 | 0.628 |
| | Person | 0.569 | 0.379 | 0.771 | 0.699 |
| YOLOv5x | Car | 0.805 | 0.784 | 0.652 | 0.623 |
| | Person | 0.559 | 0.602 | 0.738 | 0.746 |
| YOLOv7x | Car | 0.808 | 0.792 | 0.669 | 0.663 |
| | person | 0.561 | 0.603 | 0.731 | 0.786 |

## B. Result According to Precipitation

The RGB image according to precipitation is shown in Fig. 4 After labeling the car, person, cycle, and motorcycle in this image, we tested the mentioned models on CARLA dataset, and the results about car were shown in Fig. 8. and Fig. 9. Red line is for YOLOv7x, green line is for YOLOv5x and blue line for YOLOv4. In this graph, the AP score and F1 score gradually decreased as precipitation increases by 10mm/h in the standard normal weather. For example, the AP score of cars in Yolov7x recorded an AP score of 0.669 in normal situations and 0.46 in 100mm/h situations, showing a decrease in performance by about 31.24%. In the case of Yolov5x, the performance decreased by 30.67%, recording 0.652 in the normal situation and 0.452 in the 100mm/h situation. Finally, in the case of yolov4, the performance decreased by 33.75% from 0.64 to 0.424 in normal and 100mm/h condition respectively.

Table III and IV thoroughly showed that the AP score and F1 score were gradually decreased as the precipitation increased. Specifically, for person, the AP score was decreased by at least 30.23% to at most 42.14%, and the F1 score was also decreased by at least 21.76% to at most 34.95%. For bicycles, AP scores decreased by 25.96% from at least 24.47%, and F1 scores decreased by up to 39.19% from at least 24.08%. Finally, in the case of motorcycle, both AP score and F1 score were dramatically decreased by 99% or 100%. Moreover, we showed these experimental results that the object detection performance was clearly decreased as precipitation increased as shown in Fig. 10 and Fig. 11.

TABLE III. AP SCORE DEGRADATION OF OBJECT DETECTON PERFORMANCE ACCORDING TO PRECIPITATION.

| Index values | | AP Score | | | |
|---|---|---|---|---|---|
| | | Car | Person | Bicycle | Motor cycle |
| YOLOv4 | Normal | 0.64 | 0.771 | 0.913 | 0.263 |
| | 100mm/h | 0.424 | 0.469 | 0.677 | 0.0043 |
| YOLOv5x | Normal | 0.652 | 0.738 | 0.813 | 0.425 |
| | 100mm/h | 0.452 | 0.427 | 0.614 | 0.0031 |
| YOLOv7x | Normal | 0.669 | 0.731 | 0.936 | 0.198 |
| | 100mm/h | 0.46 | 0.51 | 0.693 | 0.00025 |

TABLE IV. F1 SCORE DEGRADATION OF OBJECTS DETECTION PERFORMANCE ACCORDING TO PRECIPITATION.

| Index values | | F1 Score | | | |
|---|---|---|---|---|---|
| | | Car | Person | Bicycle | Motor cycle |
| YOLOv4 | Normal | 0.629 | 0.699 | 0.802 | 0.179 |
| | 100mm/h | 0.477 | 0.414 | 0.627 | 0 |
| YOLOv5x | Normal | 0.623 | 0.746 | 0.724 | 0.415 |
| | 100mm/h | 0.514 | 0.485 | 0.535 | 0 |
| YOLOv7x | Normal | 0.663 | 0.786 | 0.859 | 0.159 |
| | 100mm/h | 0.533 | 0.549 | 0.522 | 0 |

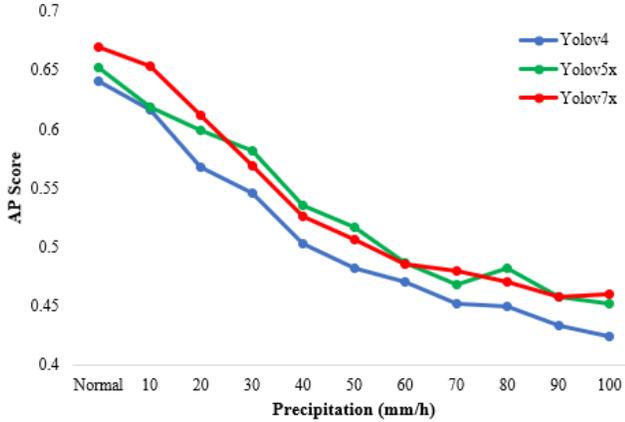

Figure 8. AP score of car according to precipitation.

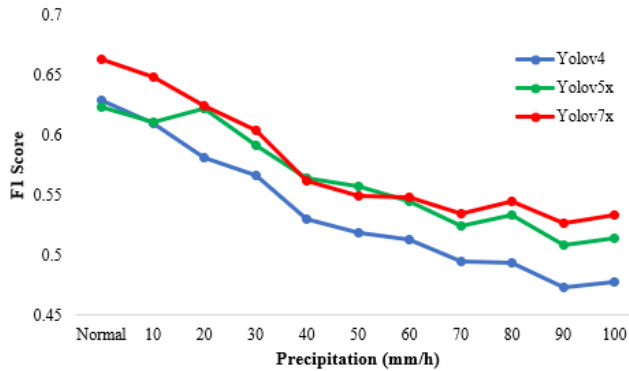

Figure 9. F1 score of car according to precipitation.

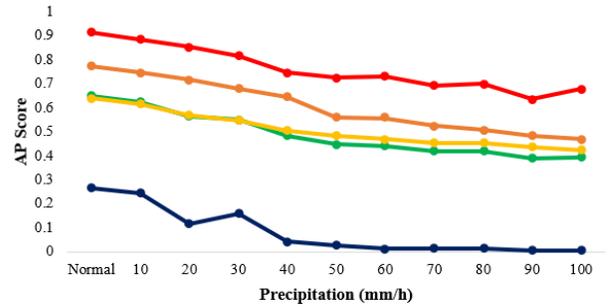

(a)

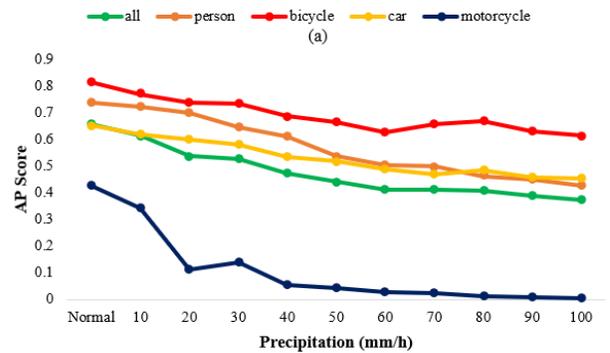

(b)

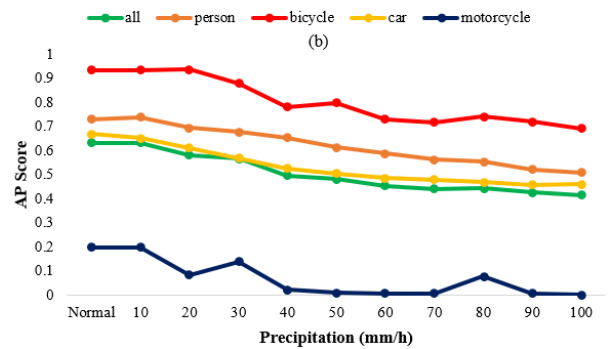

(c)

Figure 10. AP score of all instances according to precipitation: (a) YOLOv4, (b) YOLOv5x, (c) YOLOv7x.

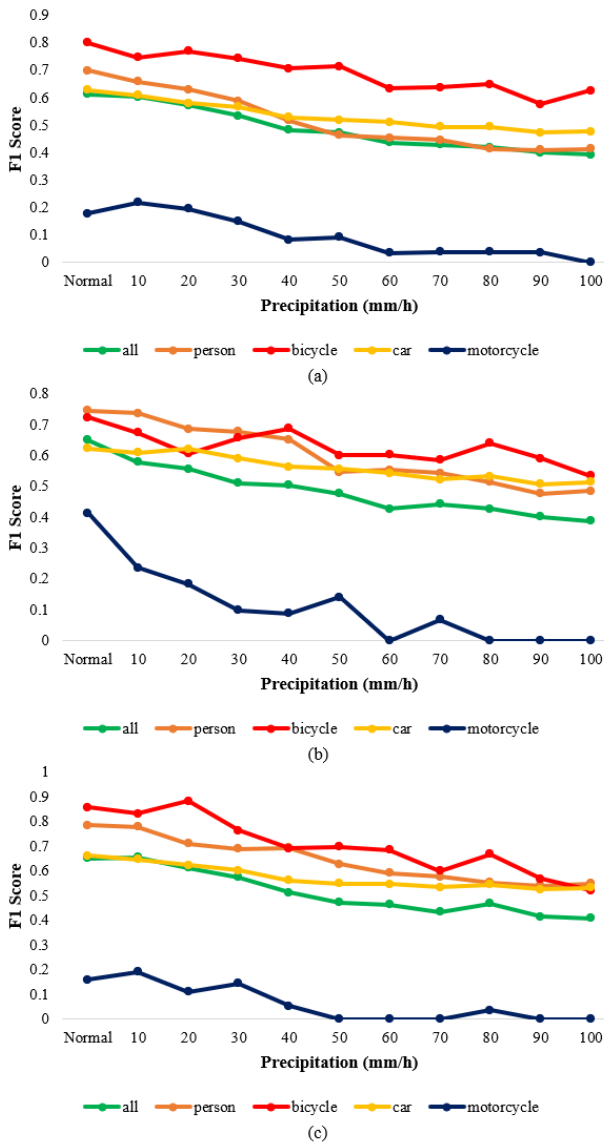

Figure 11. F1 score of all instances according to precipitation:
(a) YOLOv4, (b) YOLOv5x, (c) YOLOv7x.

Especially, in the case of a motorcycle, its performance was sharply dropped, while other objects with clearly distinguished colors were relatively detected well. This result showed that it was more difficult to detect because the background and object colors were very similar, and the whole image had only 10 instances, resulting in a rapid change in detecting process. In the case of the other three objects (i.e., car, person, and bicycle) except for the motorcycle, the color was clear, and its background was well distinguished, resulting in reasonable results even in the heavy precipitation conditions.

The detection results for a single image for YOLOv7x, which showed consistently high performance in AP Score and F1 Score, were shown in Fig. 12. In normal condition, humans and some vehicles were fairly well detected, but as precipitation increased from 0mm/h to 100mm/h, accuracy dramatically decreased from 92% to 83% and 84% to 80%. Moreover, under some precipitation conditions, bicycle recognized as motorcycles, raindrops were detected as humans.

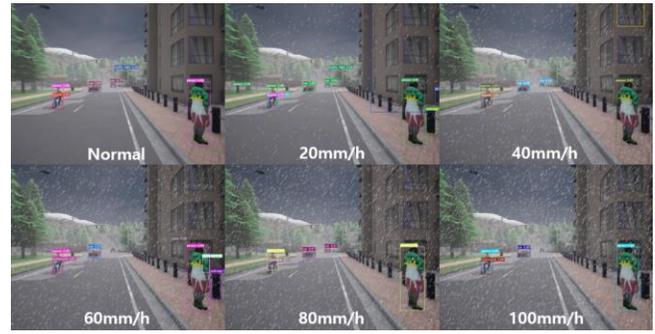

Figure 12. Result of performance degradation of object detection in diverse rain situations on YOLOv7x.

In some cases, heavy precipitation prevented detecting the vehicles which were previously recognized as vehicles

## V. CONCLUSION

In this paper, the performance of YOLOv7 with state-of-the-art and its sub-models YOLOv5 and YOLOv4 in terms of rainfall rate was thoroughly evaluated based on CARLA simulation by using synthetic rain image dataset. When comparing AP score of origin image without raindrops and AP score based on the KITTI dataset, we obtained similar performance and then could secure the reliability of simulation. Moreover, for this study, a more realistic synthetic image was collected by using experimentally obtained dataset, which were generated by using linearly extrapolated precipitation equation. As a result, it was confirmed that the performance of the object detection model was substantially degraded as precipitation increased. Consequently, when the precipitation reached 100mm/h, it was found that there was a dramatic performance degradation of more than 30% compared to the normal condition.

Moreover, in this study, we thoroughly investigated a precipitation-based object detection performance, which was not obviously conducted in other studies so far. However, even with the same precipitation, the aspect of rain can be changed due to wind or clouds. Therefore, in future, for realistic verification in an actual bad weather situation, it is necessary to implement a realistic rain situation. Furthermore, various bad weather situation simulations and evaluations are also needed. For example, it requires to verify object detection performance based on the amount of snowfall and visible distance for weather conditions such as snow, fog, and dust. Finally, we also plan to develop a sim to real dataset production and object detection network model through CARLA simulation so that it can robustly detect diverse objects even in bad weather conditions.

ACKNOWLEDGMENT

This work was supported by the Daegu Gyeongbuk Institute of Science and Technology (DGIST) Institution-Specific Project and supported by the National Research Foundation of Korea (NRF) grant funded by the Korea government (MSIT) from the Ministry of Science and Information and Communications Technology (ICT) of Korea. (No. 22-DPIC-17 and 2021R1F1A1046197).

## REFERENCES

[1] S. Ionita, "Autonomous vehicles: From paradigms to Technology," *IOP Conference Series: Materials Science and Engineering*, vol. 252, p. 012098, 2017.

[2] C. Gkartzonikas and K. Gkritza, "What have we learned? A review of stated preference and choice studies on Autonomous Vehicles," *Transportation Research Part C: Emerging Technologies*, vol. 98, pp. 323–337, 2019.

[3] D. J. Yeong, G. Velasco-Hernandez, J. Barry, and J. Walsh, "Sensor and Sensor Fusion Technology in autonomous vehicles: A Review," *Sensors*, vol. 21, no. 6, p. 2140, 2021.

[4] Y. Zhang, A. Carballo, H. Yang, and K. Takeda, "Autonomous Driving in Adverse Weaher Conditions: A Survey," *arXiv preprint arXiv:2112.08936*, 2021.

[5] A. Cord and N. Gimonet, "Detecting unfocused raindrops: In-vehicle multipurpose cameras," *IEEE Robotics & Automation Magazine*, vol. 21, no. 1, pp. 49–56, 2014.

[6] K. Garg and S. K. Nayar, "When does a camera see rain?," Tenth IEEE International Conference on Computer Vision (ICCV'05) Volume 1, 2005.

[7] S. Zang, M. Ding, D. Smith, P. Tyler, T. Rakotoarivelo, and M. A. Kaafar, "The impact of adverse weather conditions on autonomous vehicles: How rain, snow, fog, and hail affect the performance of a self-driving car," *IEEE Vehicular Technology Magazine*, vol. 14, no. 2, pp. 103–111, 2019.

[8] K. Yoneda, N. Suganuma, R. Yanase, and M. Aldibaja, "Automated Driving Recognition Technologies for adverse weather conditions," *IATSS Research*, vol. 43, no. 4, pp. 253–262, 2019.

[9] X. Fu, J. Huang, X. Ding, Y. Liao, and J. Paisley, "Clearing the skies: A deep network architecture for single-image rain removal," *IEEE Transactions on Image Processing*, vol. 26, no. 6, pp. 2944–2956, 2017.

[10] X. Li, J. Wu, Z. Lin, H. Liu, and H. Zha, "Recurrent squeeze-and-excitation context aggregation net for single image deraining," *Computer Vision – ECCV 2018*, pp. 262–277, 2018.

[11] D. Ren, W. Zuo, Q. Hu, P. Zhu, and D. Meng, "Progressive image deraining networks: A better and simpler baseline," *2019 IEEE/CVF Conference on Computer Vision and Pattern Recognition (CVPR)*, 2019.

[12] S.W. Zamir, A. Arora, S. Khan, M. Hayat, F.S. Khan, and M.H. Yang, "Restormer: Efficient transformer for high-resolution image restoration,", *Proceedings of the IEEE/CVF Conference onf Computer Vision and Pattern Recognition (CVPR)*, pp. 5728-5739, 2022.

[13] R. Walambe, A. Marathe, K. Kotecha, and G. Ghinea, "Lightweight Object Detection Ensemble Framework for autonomous vehicles in challenging weather conditions," *Computational Intelligence and Neuroscience*, vol. 2021, pp. 1–12, 2021.

[14] J. Choi, D. H. Kim, S. Lee, S. H. Lee, and B. C. Song, "Synthesized rain images for deraining algorithms," *Neurocomputing*, vol. 492, pp. 421–439, 2022.

[15] K. M. Jeong and B. C. Song, "Image synthesis algorithm for road object detection in Rainy weather," *IEIE Transactions on Smart Processing & Computing*, vol. 7, no. 5, pp. 342–349, 2018.

[16] C. Hu, S. Hudson, M. Ethier, M. Al-Sharman, D. Rayside, and W. Melek, "Sim-to-real domain adaptation for Lane Detection and classification in autonomous driving," *2022 IEEE Intelligent Vehicles Symposium (IV)*, 2022.

[17] A. Dosovitskiy, G. Ros, F. Codevilla, A. Lopez, and V. Koltun. "CARLA: An open urban driving simulator," in *Conference on robot learning*. PMLR, pp. 1-16, 2017.

[18] A. Geiger, P. Lenz, and R. Urtasun, "Are we ready for autonomous driving? The Kitti Vision Benchmark Suite," *2012 IEEE Conference on Computer Vision and Pattern Recognition*, 2012.

[19] C.Y. Wang, A. Bochkovskiy, and H.Y.M. Liao, "YOLOv7: Trainable bag-of-freebies sets new state-of-the-art for real-time object detectors," *arXiv preprint arXiv:2207.02696*, 2022.

[20] A. Bochkovskiy, C.Y. Wang, and H.Y.M. Liao, "Yolov4: Optimal speed and accuracy of object detection," *arXiv preprint arXiv:2004.10934*. 2020.

[21] Jocher Glenn. YOLOv5 release v6.1. https://github.com/ultralytics/yolov5/releases/tag/v6. 1, 2022. (yolov5)

[22] M. Hnewa and H. Radha, "Object detection under rainy conditions for autonomous vehicles: A review of state-of-the-art and emerging techniques," *IEEE Signal Processing Magazine*, vol. 38, no. 1, pp. 53–67, 2021.

[23] F. Yu, W. Xian, Y. Chen, F. Liu, M. Liao, V. Madhavan, and T. Darrell, "Bdd100k: A diverse driving video database with scalable annotation tooling," *arXiv preprint arXiv:1805.04687*. 2(5), 6, 2018.

[24] H. Zhang, L. Xiao, X. Cao, and H. Foroosh, "Multiple adverse weather conditions adaptation for object detection via causal intervention," *IEEE Transactions on Pattern Analysis and Machine Intelligence*, pp. 1–1, 2022.

[25] W. Liu, G. Ren, R. Yu, S. Guo, J. Zhu, and L. Zhang, "Image-adaptive YOLO for object detection in adverse weather conditions," *Proceedings of the AAAI Conference on Artificial Intelligence*, vol. 36, no. 2, pp. 1792–1800, 2022.

[26] R. Walambe, A. Marathe, K. Kotecha, and G. Ghinea, "Lightweight Object Detection Ensemble Framework for autonomous vehicles in challenging weather conditions," *Computational Intelligence and Neuroscience*, vol. 2021, pp. 1–12, 2021.

[27] T. Sharma, B. Debaque, N. Duclos, A. Chehri, B. Kinder, and P. Fortier, "Deep learning-based object detection and scene perception under bad weather conditions," *Electronics*, vol. 11, no. 4, p. 563, 2022.

[28] X.-Z. Chen, C.-M. Chang, C.-W. Yu, and Y.-L. Chen, "A real-time vehicle detection system under various bad weather conditions based on a deep learning model without retraining," *Sensors*, vol. 20, no. 20, p. 5731, 2020.

[29] A. M. ABDULGHANİ and G. G. MENEKŞE DALVEREN, "Moving object detection in video with algorithms Yolo and faster R-CNN in different conditions," *European Journal of Science and Technology*, 2022.

[30] H. Jeon, Y. O. Kim, M. Choi, D. Park, S. Son, J. Lee, G. Choi, and Y. Lim, "Carla Simulator-based evaluation framework development of Lane Detection Accuracy Performance under sensor blockage caused by heavy rain for Autonomous Vehicle," *IEEE Robotics and Automation Letters*, vol. 7, no. 4, pp. 9977–9984, 2022.

[31] Z. Wang, A. C. Bovik, H. R. Sheikh, and E. P. Simoncelli, "Image quality assessment: From error visibility to structural similarity," *IEEE Transactions on Image Processing*, vol. 13, no. 4, pp. 600–612, 2004.

[32] A. Hore and D. Ziou, "Image quality metrics: PSNR vs. SSIM," *2010 20th International Conference on Pattern Recognition*, 2010.